 \documentclass[sigconf, nonacm]{acmart}
 
\usepackage{multirow}
\usepackage{stfloats}
\AtBeginDocument{%
  \providecommand\BibTeX{{%
    \normalfont B\kern-0.5em{\scshape i\kern-0.25em b}\kern-0.8em\TeX}}}

\begin{document}

\setcopyright{acmcopyright}
\copyrightyear{2023}
\acmYear{2023}

\acmConference[Long Beach '23]{Long Beach '23: ACM SIGKDD Conference on Knowledge Discovery \& Data Mining}{August 06--10, 2023}{Long Beach, CA}


\title[Seismic Fatality Estimation from Crowdsourced Data based on LLMs]{Near-real-time Earthquake-induced Fatality Estimation using Crowdsourced Data and  Large-Language Models}


\author{Chenguang Wang *}
\affiliation{%
  \institution{Stony Brook University}
  \streetaddress{100 Nicolls Road}
  \city{Stony Brook}
  \state{New York}
  \country{USA}
  \postcode{11794}
  }

\author{Davis Engler *}
\affiliation{%
  \institution{U.S. Geological Survey}
  \city{Golden}
  \state{Colorado}
  \country{USA}
  }
  
\author{Xuechun Li}
\affiliation{%
  \institution{Johns Hopkins University}
  \streetaddress{3400 N. Charles Rd}
  \city{Baltimore}
  \state{Maryland}
  \country{USA}
  \postcode{21218}
  }
\author{James Hou}
\affiliation{%
  \institution{Stony Brook University}
  \streetaddress{100 Nicolls Road}
  \city{Stony Brook}
  \state{New York}
  \country{USA}
  \postcode{11794}
  }

\author{David J. Wald}
\affiliation{%
  \institution{U.S. Geological Survey}
  \city{Golden}
  \state{Colorado}
  \country{USA}
  }
\author{Kishor Jaiswal}
\affiliation{%
  \institution{U.S. Geological Survey}
  \city{Golden}
  \state{Colorado}
  \country{USA}
  }
\author{Susu Xu}
\affiliation{%
  \institution{Johns Hopkins University}
  \streetaddress{3400 N. Charles Rd}
  \city{Baltimore}
  \state{Maryland}
  \country{USA}
  \postcode{21218}
}

\begin{abstract}
When a damaging earthquake occurs, immediate information about casualties (e.g., fatalities and injuries) is critical for time-sensitive decision-making by emergency response and aid agencies in the first hours and days.  Systems such as Prompt Assessment of Global Earthquakes for Response (PAGER) by the U.S. Geological Survey (USGS) were developed to provide a forecast of such impacts within about 30 minutes of any significant earthquake globally. 
However, existing disaster-induced human loss estimation systems often rely on early casualty reports manually retrieved from global traditional media, which are labor-intensive, time-consuming, and have significant time latencies. Recent approaches use keyword matching and topic modeling to identify human casualty-relevant information from social media, but tend to be error-prone when dealing with complex semantics in multi-lingual text data, and parsing dynamically changing and conflicting human death and injury number shared by various unvetted sources in social media platforms. 

In this work, we introduce an end-to-end framework to significantly improve the timeliness and accuracy of global earthquake-induced human loss forecasting using multi-lingual, crowdsourced social media. Our framework integrates  (1) a hierarchical casualty extraction model built upon large language models, prompt design, and few-shot learning to retrieve quantitative human loss claims from social media,  (2) a physical constraint-aware, dynamic-truth discovery model that discovers the truthful human loss from massive noisy and potentially conflicting human loss claims, and (3) a Bayesian updating loss projection model that dynamically updates the final loss estimation using discovered truths. We test the framework in real-time on a series of global earthquake events in 2021 and 2022 and show that our framework effectively automates the retrieval of casualty information faster but with comparable accuracy to those now retrieved manually by the USGS.

\end{abstract}

\vspace{-1.5cm}

\setlength{\belowdisplayskip}{0pt} \setlength{\belowdisplayshortskip}{0pt}
\setlength{\abovedisplayskip}{0pt} \setlength{\abovedisplayshortskip}{0pt}

\maketitle
{
\renewcommand{\thefootnote}{\fnsymbol{footnote}}
\footnotetext[1]{Equal Contributions}
}
\vspace{-0.2cm}
\section{Introduction}
\label{intro}
\begin{figure*}[!t]
\vspace{-0.3cm}
\centerline{\includegraphics[width=1.7\columnwidth]{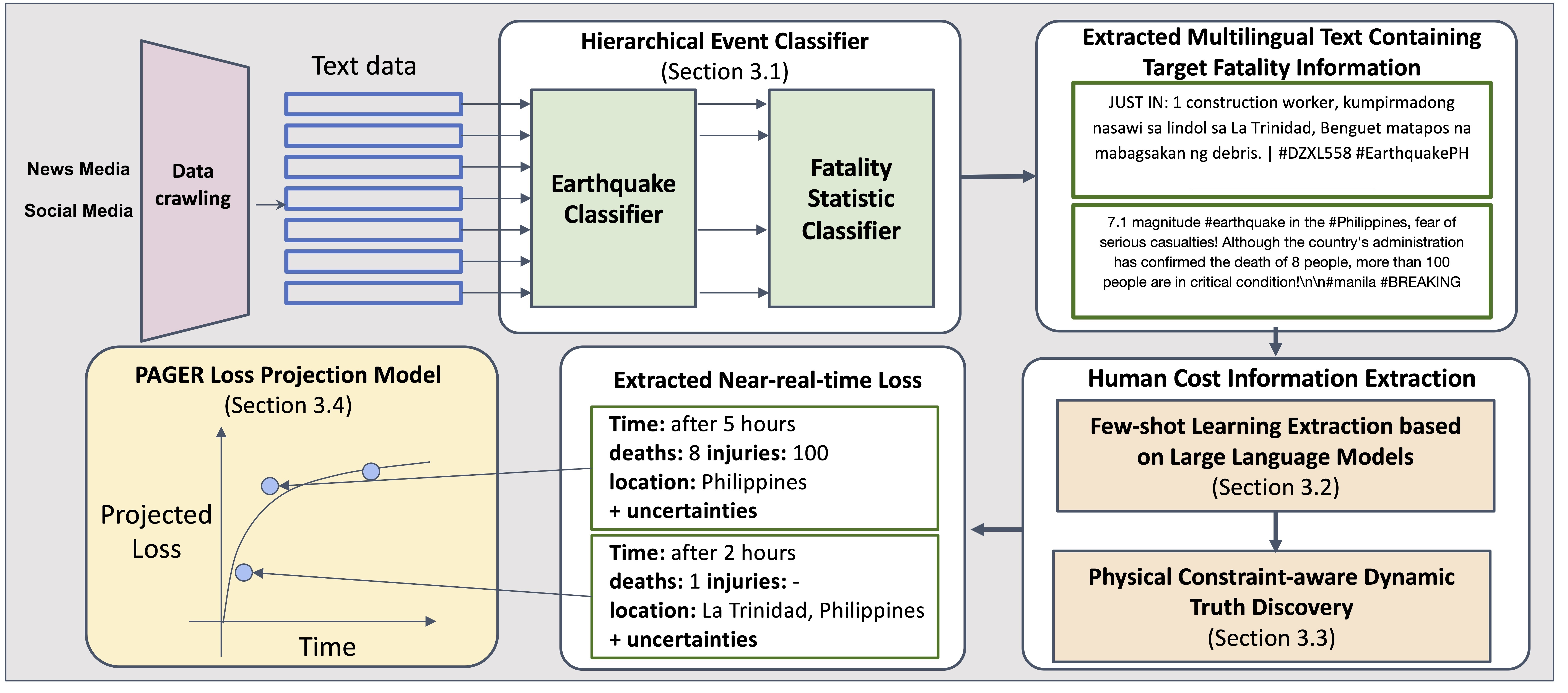}}
\caption{Overview of our framework's design and application. Texts are crawled using keyword searches from social media and news sources. The data are first filtered with Hierarchical Event Classifier to extract texts that are highly possibly related to seismic human fatality (Section 3.1). Structured data (human fatality number and injury number) are further extracted from these filtered texts (Section 3.2). Furthermore, a physical constraint-aware dynamic truth discovery algorithm is introduced to cross-validate and discover the ground truth human fatality/injury number from various claims by various sources/accounts (Section 3.3). The output is eventually incorporated into an earthquake loss prediction model built on Bayesian updating to project the total human loss induced by an earthquake (Section 3.4).}
\vspace{-0.5cm}
\label{fig0}
\end{figure*}
%
Short-notice disastrous events, such as earthquakes, often cause considerable tremendous human costs, including fatalities, injuries, and displaced persons~\cite{intro1}. Immediate information concerning casualties after such natural disasters plays an important but challenging role in the post-disaster response. Traditional casualty reports are often obtained from emergency response and rescue teams in the field, which often takes weeks to months~\cite{wyss2017report,noh2020efficient}. In the interim, many global emergency aid agencies and first responders currently refer to the casualty estimates provided by the Prompt Assessment of Global Earthquakes for Response (PAGER) system, developed by the U.S. Geological Survey (USGS)~\cite{jaiswal2010development}. The PAGER reports provide a range of possible human fatalities (and economic impacts) within 30 minutes of any significant global earthquake, by updating an empirical fatality model using early casualty and injury reports and updated ground shaking maps~\cite{noh2020efficient}. However, in current practices,  these early reports of human deaths and injuries are often manually retrieved from traditional media like Reuters or CNN, which is labour-intensive and has significant time latency.

Compared to traditional information sources with time delays, social media platforms provide access for the masses to directly share their feelings and observations concerning an evolving disaster, thereby providing potentially timely and useful onsite data compared to traditional media and field surveys~\cite{houston2015social,gao2011harnessing}. For example, we found that the first social media post reporting human deaths in 2022 M7.6 Papua New Guinea Earthquake is from a personal account within the macroseismic zone, posting a snapshot of a local community Facebook forum in Wau, Papua New Guinea, indicating 3 reported deaths. 
Existing social media scraping approaches mainly focus on categorizing the relevance level of text data instead of extracting exact casualty reports (e.g., death number, injuries number) and their locations. Researchers explored classic machine learning techniques, such as Support Vector Machines, Convolutional Neural Nets, and logistic regression, in combination with pre-trained disaster/social media post word embeddings to categorize relevant information \cite{imran2013practical, ahadzadeh2021earthquake, stowe2016identifying,wang2018social}. For example, CrisisNLP is a crisis informatics effort that leverages social media to collect disaster-related Twitter data and uses classifiers with traditional topic modeling~\cite{alrashdi2019deep}. However, these approaches mainly focus on categorizing the gathered articles or text data, instead of extracting exact numbers and locations. In addition, they are not robust against the highly complex and noisy social media text with large amounts of misinformation and ambiguities, due to the limited capability of traditional word embedding and topic modeling methods.
Besides, previous work mainly focuses on English and Chinese data, neglecting the abundance of multilingual data present in global earthquakes. 

To fill these gaps, our objective is to achieve automatic retrieval of the exact number of earthquake-induced human losses (death and injury) from multi-sourced social media and traditional media platforms for global events. We identify three important challenges posed by our objective. First, the multi-lingual text data shared by people around the world often contain a variety of complex semantics. For example, use of abbreviations and jargons~\cite{imran2015processing, han2013lexical}, or recollections of casualties from past events (e.g., recall of the 2010 Haiti when searching the 2021 Haiti Earthquake) and co-occurring unrelated non-earthquake emergency events (e.g. COVID). Second, the text data from different sources often contain incorrect and potentially conflicting information from a large number of unvetted sources. For example, a piece of misinformation on social media, saying 16 deaths in one hour after 2022 M7.6 Papua New Guinea Earthquake, has been widely circulated by many verified public media accounts but was later claimed to be misinformation. The unknown reliability of various data sources makes it challenging to extract accurate information.
Third, the reported human costs dynamically evolve with heterogeneous region-specific patterns tied to resource availability, meaning the ground-truth value is changing as well. However, information spreading on social networks often takes time and exhibits delay patterns, thus, delayed data often appear more prominently than the latest, more accurate data. This largely constrains the timeliness of information retrieval and poses additional difficulties when cross-sourcing information for verification. Moreover, due to time sensitivity, it is impossible to have experts label large amounts of extracted text data for fine-tuning and adapting the information retrieval models to data reporting patterns specific to the earthquake of interest.

To address these challenges, we introduce a novel, near-real-time, end-to-end framework that can automatically retrieve accurate human casualty information from multiple data sources and adaptively by integrating Large Language Models (LLMs) and dynamic truth discovery, as shown in Figure~\ref{fig0}. Specifically, this work makes the following contributions:

\noindent\textbf{(1)} We develop a hierarchical event-specific disaster data extraction framework that leverages a multilingual event classifier, prior knowledge of LLMs, specially designed prompts, Few-Shot Learning, and dynamic truth discovery to extract exact human casualty statistics from crowdsourced text data with complex semantics, without additional training or fine-tuning. To the best of our knowledge, this is the first disaster human fatality information retrieval framework built on LLMs. 
    
    
\noindent\textbf{(2)} We design a physical constraint-aware dynamic truth discovery scheme to accurately uncover reported fatalities from noisy, incomplete, and conflicting information by considering (i) physical rules that human losses will not decrease with time, and (ii) historical reliability of different information sources. 
    
    
\noindent\textbf{(3)} We integrate the data pipeline, information extraction, and truth discovery with existing PAGER fatality loss models and enable automatic updating of the PAGER system in near-real-time seismic loss projetion, for the first time. 
    
\noindent\textbf{(4)} We evaluate and characterize the framework using three recent real-world earthquakes. The evaluation results demonstrate significant performance gain achieved by our framework, providing timely and accurate human fatality information with finer time resolution  compared to traditional approaches.
    
    

\vspace{-0.3cm}
\section{Related Work}


Although various near-real-time disaster information platforms are open-source for disaster response, there is still no framework openly available to support automatic and multi-source information retrieval and impact estimation in near real-time. The USGS PAGER system provides rapid estimates of economic losses and human fatalities~\cite{jaiswal2010development} relying on empirical models and geospatial data, and it can be updated by manually searching news sources for casualty reports \cite{noh2020efficient}. 
In the Natural Language Processing (NLP) community, many approaches have been developed to acquire disaster damage information from social media platforms like Twitter. Some studies design pre-defined keyword lists or tables to search and extract useful information from social media textual posts. After creating keyword lists for each subcategory, the keyword search-based method can identify and categorize qualified posts and contain comprehensive situational knowledge\cite{hao2020leveraging}. For example, ~\cite{deng2016new,hao2020leveraging} split information on disaster damage into multiple groups like infrastructure destruction, supply chain demands, and affected activities.
Generally, methods of keyword searching usually discover certain information from the social media text corpus. After keyword lists are created for each group, this approach can identify and categorize qualified posts. However, it is costly and time-consuming to enumerate every possible keyword and phrase related to a topic due to the colloquial nature of social media textual messages \cite{temnikova2015emterms,imran2013practical}. 
Previous work like \cite{imran2013practical, stowe2016identifying, wang2019rapid} apply existing word embeddings or train disaster social-media-post-specific word embeddings to obtain social media data representations. Afterward, machine learning methods like linear classifiers, logistic regression, and Support Vector Machines (SVMs) are fine-tuned upon the embeddings of the datasets to recognize and categorize Twitter messages. Due to the excellent performance in image classification, Convolutional Neural Networks (CNNs) are deployed extensively \cite{imran2017robust, firoj_ACL_2018embaddings, ahadzadeh2021earthquake, caragea2011classifying}. However, these methods are often ineffective when applied to unobserved events and need fine-tuning.


Transformer-based language models have recently become state-of-the-art due to their powerful attention mechanism that models inter-token relations \cite{vaswani2017attention}. The transformer models are usually trained on large amounts of online texts, making them applicable to many language tasks. Afterward, a Bidirectional Encoder Representations from Transformers (BERT) model, an Encoder-only variant of Transformers that outputs word and sentence-level representations, is applied as proposed by \cite{devlin2018bert}. BERT outperforms traditional word embedding methods in many natural language understanding tasks by providing context-aware representations.  In this study, we use notable BERT variants, RoBERTa, and XLM-RoBERTa \cite{liu2019roberta}. 

\noindent\textbf{Dynamic Truth Discovery:} 
The truth discovery problem was first formally formulated and resolved by a Bayesian heuristic algorithm, Truth Finder, in~\cite{yin2007truth}. Given estimated source weights and interactions among different claims, the confidence score of each claim is updated using Bayesian updating. Based on Truth Finder, extended models were further introduced to integrate prior knowledge, including constraints on truth patterns and source dependencies, to improve the accuracy and efficiency of truth discovery ~\cite{pasternack2010knowing,dong2009integrating}. However, significant knowledge gaps exist in finding truth from widely spread disingenuous posts and information with severe time latency under dynamically changing truths, especially for time-sensitive tasks like ours, which is a challenging task that yet has not been well addressed. 
\vspace{-0.3cm}
\section{Framework Design}
In this section, we present our framework (shown in Figure \ref{fig0}) to extract casualty data from crowdsourced reports. Specifically, this framework includes our key components: (1) an automatic data crawling pipeline that automatically scrapes data from multiple sources, (2) a hierarchical human cost value extraction module integrating a hierarchical event classifier that filters out text relevant to target earthquake events and casualty statistics (Section~\ref{hec}), and a fatality value extractor built based on LLMs and Few-Shot Learning (Section~\ref{llm}), (3) a physical constraint-aware dynamic truth discovery model that recovers casualty estimates from massive noisy and potentially conflicting data, constrained by physical rules of evolving reported fatalities (Section~\ref{dtd}), and (4) a PAGER loss-projection model that dynamically updates final human cost estimations (Section~\ref{bayesian}).
To enable near real-time disaster data retrieval, we build an event-triggering pipeline that retrieves and processes real-time disaster data, mainly text, from Twitter and News API. The pipeline enables automatic keywords and query generation as well as streaming data collection and storage. We mainly focus on extracting human casualty statistics from social media posts and news headlines and articles. 

\vspace{-0.5cm}
\subsection{Hierarchical Event Classifier} \label{hec}
Hierarchical event classifiers filter out irrelevant text data from large amounts of crowdsourced data to improve the computational efficiency of quantitative human loss data pairing. 
Our hierarchical event classifier contains two modules: a earthquake event classifier to tell if a text is relevant to a target earthquake event, and a fatality statistics classifier to determines if the text includes casualty statistics. 
The design of the integrative hierarchical event classifier is based on our observations concerning disaster text data. Two common phenomena we discovered are that (1) because disaster zones are often large in extent, there are often fatality reports therein that are not induced by the event of interest, but rather by unrelated occurrences (e.g., car accidents or pandemic); and (2) because seismic impacts are often complex, a large amount of earthquake-related information does not contain casualty statistics. Based on the observations, we design the hierarchical event classifier that cross-classifies the input text data to cull irrelevant information.  

To deal with complex semantics in multi-lingual, multi-sourced data, the two modules share the same model architecture back-bone as XLM-RoBERTa, a  state-of-the-art, cross-lingual word-embedding model and contains 350 million parameters \cite{conneau2019unsupervised} for effective word embedding. The word representations are further input to a neural network to classify if a text is relevant to an earthquake event as well as if the text contains any casualty statistics. XLM-RoBERTa was pre-trained on text spanning 100 languages, giving it multilingual understanding and cross-lingual transfer. The cross-lingual transfer capability enables us to only train the models with the abundant English language and generalize our classification to more resource-scarce languages.
We further train the earthquake classifier and fatality statistics classifier separately using labeled disaster corpus, CrisisNLP \cite{imran2016lrec}. CrisisNLP labels them through crowd-sourcing efforts. The social media posts enclose various disaster events (e.g., earthquakes, hurricanes, pandemics), labels that describe whether a social media post is relevant to the disaster, and statistics. 
\vspace{-0.3cm}
\subsection{Human Loss Extraction via LLMs}\label{llm}

With most irrelevant information filtered out by the hierarchical event classifier, we further extract the casualty numbers. The human casualty extractor plays two important roles: (1) to extract the exact number of casualties, including fatalities, injuries, and locations, and (2) to provide second verification of the relevance of the text data, e.g., past earthquake occurrences recalled in the same region, which can not be directly differentiated by the event classifier.
To our best knowledge, ours is the first study to accomplish near-real-time casualty value extraction on crowdsourced text data other than the classification of messages. The desired details are often embedded within crowdsourced text data with complex language. 
For example, issues such as irregular syntax, use of the conjunctions 'and' or 'or,' abbreviations, and confusing numerical expressions need to be addressed. For example, when searching the 2021 M7.2 Haiti earthquake in Twitter, a relevant post is as follows:
\begin{figure}[htbp]
\hspace{-1mm}\begin{center}
\scalebox{0.9}{
\begin{tabular}{c}
\includegraphics[width=0.4\textwidth]{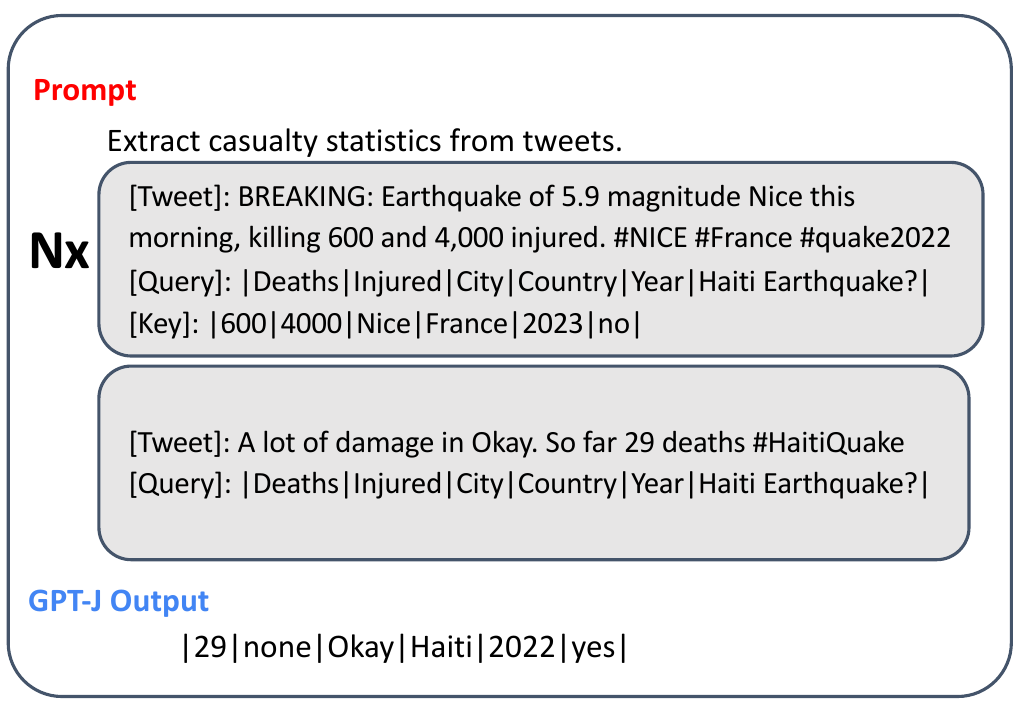}
\end{tabular}}
\vspace{-0.3cm}
\caption{A conceptual diagram of our Few-Shot Learning prompt approach to extracting information. Nx represents the number of examples (Shots) that we give in the prompt. }
\vspace{-0.5cm}
\label{fig:prompt}
\end{center}
\end{figure}

\textit{``8/21 Haiti was hit by an earthquake leaving 2,200 dead, 10K homeless. 1 week later a Hurricane, killing 14, caused 500mil in damage. 1 month b4 they're Pres was killed leaving the isle lawless. Those are refugees fleeing death \&amp; devastation, they have nothing left to go back to."}

\begin{figure*}[h]
\begin{center}
\scalebox{1}{
\begin{tabular}{c}
\includegraphics[width=2\columnwidth]{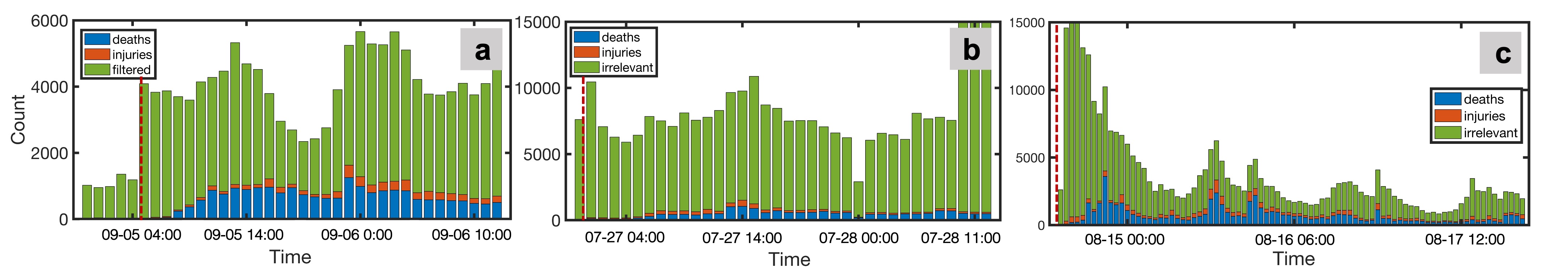}
\vspace{-0.5cm}
\end{tabular}}
\caption{Distributions of social media posts filtered for relevance, posts mentioning fatalities, and posts mentioning injuries obtained for (a) the 2022 Luding, China earthquake, (b) the 2022 Philippines earthquake,  (a) the 2021 Haiti earthquake as time evolves (UTC time). The red line represents when the mainshock occurred.}
\vspace{-0.5cm}
\label{fig:count}
\end{center}
\end{figure*}

This tweet contains multiple quantitative values related to the targeted earthquake (2,200 dead and 10,000 homeless), but also irrelevant information about a hurricane one week later than the earthquake (14 death, 500 million damage). 
Moreover, many multi-lingual abbreviations, such as local city names, cannot be directly filtered out by traditional rule-based methods. For example, we found this post for the aforementioned Haiti earthquake:

\textit{`` A lot of damage in Okay. So far 29 reported deaths. "}

``Okay" is Haitian Creole for Les Cayes, a major port and city in Haiti that suffered severe damage during the 2021 Haiti earthquake.
Because our system targets global earthquakes, it is impossible for the traditional natural language processing techniques—such as imposing manually designed rule-based or keyword matching—to handle such highly flexible text data reporting patterns varying with local social-cultural characteristics.

To address these challenges, we hereby use the new generation of LLMs and Few-Shot Learning to conduct unstructured data pairing for accurate and efficient casualty data extraction from crowdsourced text data.
\vspace{-0.3cm}
\subsubsection{Backbone LLMs}  Among the commonly used transformer-based language models, the Generative Pre-trained Transformer (GPT), introduced by \cite{radford2018improving}, achieves the most robust text generation. The GPT is an autoregressive model that uses seed text as context to generate new text~\cite{radford2019language,brown2020language}. 
GPT-2 (1.5 billion parameters)~\cite{radford2019language} and GPT-3 (175 billion parameters)~\cite{brown2020language} are developed to enhance the capability of the model to recognize patterns present within the input text, without the need for fine-tuning using fully labeled datasets.
In this work, we utilize GPT-J, an open-source alternative to the GPT family \cite{gpt-j}. 
Although GPT-J only has 6 billion parameters, the model sufficiently retains the capabilities and embedded knowledge present within the comparably larger GPT-3 (175 billion). 

Because casualty estimation frameworks like the PAGER system are particularly sensitive, the LLM doubles as a second layer of defense. We can eliminate distracting information by utilizing extracted information such as location and earthquake specificity. If we extract a statistic but cannot convert it to a number, we will discard it. Although this method is generally reliable, generative models can still produce random or inexact answers. These errors come from close numbers, random characters, and related words. To combat this, we run a beam search for the most likely response at inference. We further assure the data reliability by tracking the uncertainties of each produced token and limiting the probabilities to a specific range. 


\vspace{-0.3cm}
\subsubsection{Prompt Design} 
Few-Shot Learning in natural language processing mainly refers to the practice of feeding a pre-trained language model with a very small number of natural language templates, i.e., \textbf{"prompt,"} as opposed to fine-tuning methods that require a large amount of training data ~\cite{shin2020autoprompt}. It helps the model adapt to the desired task with decent accuracy. This technique enables the model to generalize to understand related but previously unseen tasks with just a few examples.
One of the key elements of Few-Shot Learning is prompt design. 
In our prompt design, we attach data to each text and ask the model to replicate the examples. Each example contains one or two sentences of text, a query, and the responses to the queries as follows:

\noindent\textit{[Tweet]: BREAKING: Earthquake of 5.9 magnitude in Nice this morning, killing 600 and 4k injured. \#France\#NICE}

\noindent\textit{[Query]:deaths|injuries|location|Cities|Country|Year|Haiti Earthquake?}

\noindent\textit{[Key]:600|4000|Nice|Nice|France|2021|No})

We include examples that cover possible edge cases and missing information for increased robustness. For instance, some example text data do not contain injury statistics, and we will replace the response with a unique character that designates it as missing. We encourage the model to fill in incomplete or obfuscated information like location with its pre-existing knowledge base from the LLMs (e.g., recognizing that Okay is Haitian Creole for Les Cayes and that it is a city in Haiti).


\vspace{-0.3cm}
\subsection{Dynamic Truth Discovery}\label{dtd}
\vspace{-0.1cm}

The truth discovery problem was first introduced to find the true claim from multiple claims shared by different information resources~\cite{yin2007truth}. Researchers proposed multiple models (e.g., AVGLog, Invest, and PooledInvest) to handle the source dependency and heterogeneous source credibility in truth discovery problems~\cite{dong2009integrating, pasternack2010knowing}. However, finding the truth when many posts are disingenuous and ground truth dynamically changes is still a challenging task.
In this work, dynamic truth discovery is designed to integrate multiple different information sources to yield a distribution of casualty values $p_t=(p_t^1, \cdots, p_t^k, \cdots, p_t^K)$. 
To explicitly model the quality of estimations from different sources, we design an information score ($IS_{i,t}^k$) to quantify the contributions of each data source $i$ to the belief of human cost value $k$ at a certain time point $t$. The information score is designed based on three aspects of the information credibility from a specific data source $i$: confidence score ($\xi_{i,t}^{u,k}$), relevance score ($r_{i,t}^{u,k}$), and independence score ($\rho_{i,t}^{u,k}$):

\noindent\emph{Confidence score ($\xi_{i,t}^{u,k}$)} quantifies the confidence level of the extracted human cost variable $k$ from a text data point $u$ provided by the source $i$, with a range of $(0,1)$.  This score can be obtained from the confidence level of a large language model when answering a fatality query. 

\noindent\emph{Relevance score ($r_{i,t}^{u,k}$)} measures if a text data $u$ output value of $k$ is relevant to casualty information in the target disaster event, with a range of $(0,1)$.  Relevance is obtained by the probability output from the hierarchical event classifier and the probability of LLM's answers to query questions about the event. 

\noindent\emph{Independence score ($\rho_{i,t}^{u,k}$)} depicts if a text data $u$ indicating casualties $k$ is original data instead of forwarding/copying information from other earlier text data, with a range of $(0,1)$.  This score is obtained based on if a post is significantly similar to an earlier post or cites information from another source. Higher scores mean that the data source is more independent.
Integrating the above scores, we define the information score as 
\begin{align*}
IS_{i,t}^k = \sum_u \xi_{i,t}^{u,k}*r_{i,t}^{u,k}*\rho_{i,t}^{u,k}.
\end{align*}
By normalizing the score across multiple individual accounts $i\in I$ to a range of $[0,1]$, we get a normalized information score $NIS_{i,t}^k$. We also impose physical constraints to further calibrate the information score. The physical constraint is based on order statistics, i.e., \emph{that fatality numbers should not decrease with time}. Therefore, the transitions from $p_{t-1}$ to $p_t$ should be subject to a constrained transition matrix, i.e., an upper triangular matrix, due to the probability of transiting from value $m$ to any $n<m$ is zero. We can further obtain a hard upper bound for each value $k$'s probability, where $k\leq K$,  at time point $t$:
\[NIS_{i,t}^k \leq p_t^k\leq \max (p_{t-1}^1,\cdots,p_{t-1}^k ).\] For example, if at time point $t-1$, the probability of fatalities value $0$ is $0$, then the probability that it will become $0$ at time point $t$ is $0$, because the number of deaths will only stay the same or increase.
To impose this physical constraint, we will prohibit the invalid transition by removing the corresponding $IS_{i,t}^k$. 

We also design a source reliability score ($s_{i}$) to quantify the reliability of the information provided by the source $i$. We denote source $i$'s output set as $g(i)$ and the set of sources that can output value $k$ as $f(k)$. The reliability score is measured based on historical information credibility by summing up all the information scores of the source $i$. We apply a sigmoid function to normalize its scale between 0 and 1 and get $D_t^k$. 
\begin{equation}
\begin{aligned}
s_i = \dfrac{ \sum_{k\in g(i),t}I(IS_{i,t}^k)D_t^k+(1-I(IS_{i,t}^k))(1-D_t^k)}{\sum_{k\in g(i),t}|IS_{i,t}^k|},
\end{aligned}
\end{equation}
where $D_t^k = \dfrac{1}{1+ \exp(-\sum_{i\in f(k)} IS_{i,t}^k)}.$ $I(x)$ is an indicator function in which $I(x) = 1$ when $x>0$ or else it is $0$. The reliability score evaluates the ability of a source to provide high-fidelity estimates agreed by other high-fidelity sources. We finally obtain the updated probability distribution of the values as
\begin{equation}
\begin{aligned}
p_t^k = \dfrac{\sum_{i\in g(k)}s_iNIS_{i,t}^k}{\sum_{i\in g(k),k \in f(i)}s_iNIS_{i,t}^k}.
\end{aligned}
\end{equation}
To ensure the physical constraints persist, we will take the upper bound of $p_t^k$ if the value is higher than the upper bound. By fusing different sources, the aggregated estimate at time $t$ is:
\begin{equation}
\begin{aligned}
k^*_t = arg\max_k p_t^k.
\end{aligned}
\end{equation}
The physical constraints-aware, dynamic truth discovery scheme finally outputs casualty values hourly, mainly deaths and injuries for the target event in each target country. We further use the timestamp of the first text reporting the corresponding human cost value, as the corresponding time label to obtain a reliable time-series of casualties devoid of any duplications or redundancy.
\begin{figure}[htbp]
\hspace{-1mm}\begin{center}
\scalebox{0.9}{
\begin{tabular}{c}
\includegraphics[width=0.5\textwidth]{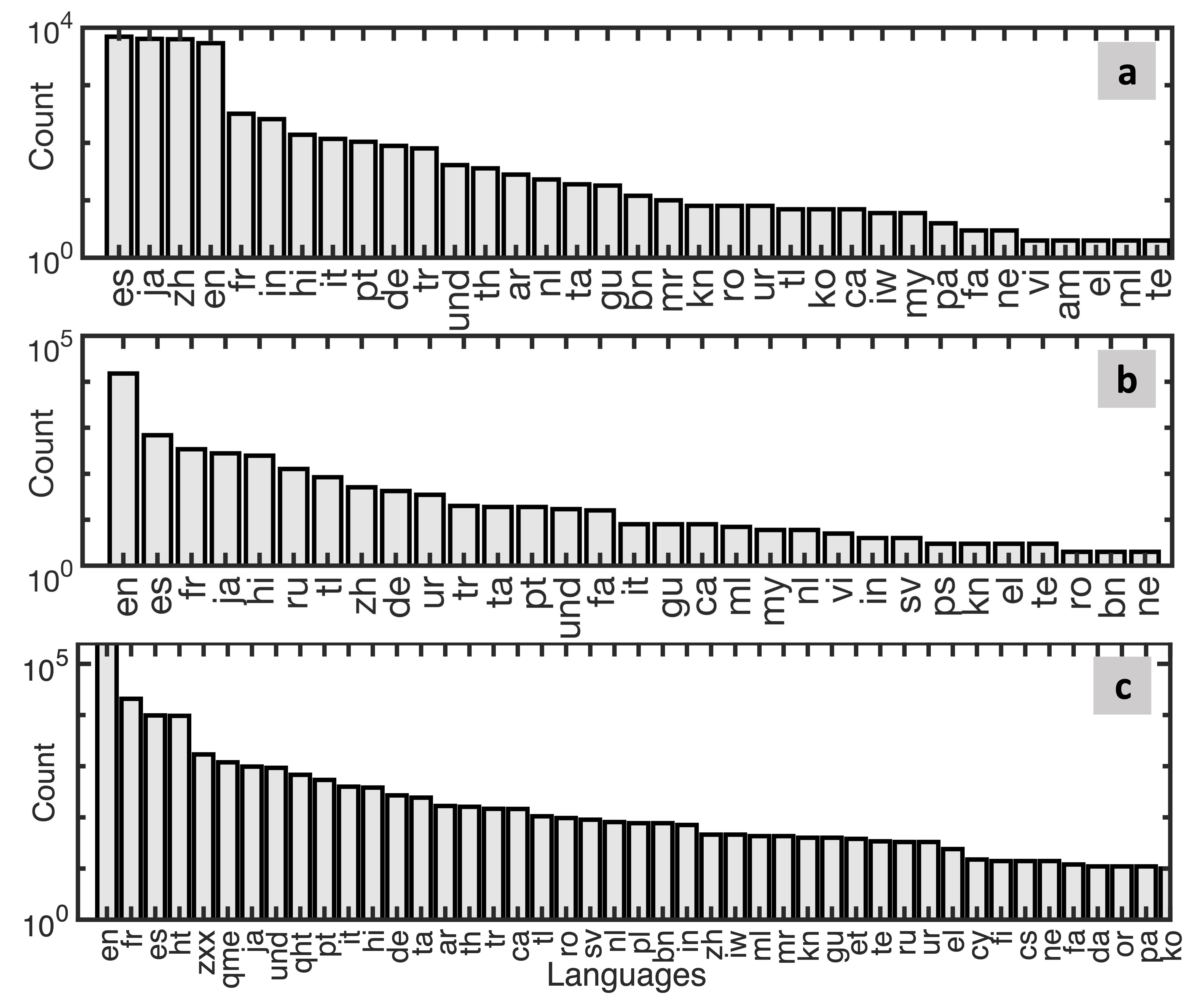}
\end{tabular}}
\vspace{-0.3cm}
\caption{Language distribution of Twitter data retrieved for different earthquake events: (a) the 2022 Luding, China earthquake, (b) the 2022 Philippines earthquake, (c) the 2021 Haiti earthquake.}
\label{fig:language}
\end{center}
\vspace{-0.6cm}
\end{figure}
\vspace{-0.3cm}
\subsection{Fatality Estimate Projections }\label{bayesian}
Past studies show that reported losses for many earthquakes follow a simple (but initially unpredictable) exponential cumulative distribution function, determined by parameter $\alpha$. The loss projection model can be formulated as 
\begin{equation}
\begin{aligned}
N(t) = N_{\infty}(1-\exp(-\alpha t))
\end{aligned}
\end{equation}
With fatalities reports extracted from crowdsourced data, we can update our estimates of the parameter $\alpha$ utilizing Bayesian updating. In this work, we follow the Bayesian updating algorithm used by the current PAGER system~\cite{noh2020efficient} to enable efficient fatality projection updating. The approach integrates the uncertainties of new observations from reported data with the {\it a priori} model learned from historical events occurring in similar regions. Currently, due to the significant impacts on the PAGER system results, USGS experts still need to carefully review and validate the aggregated fatality estimates extracted from dynamic truth discovery before integrating them for human loss projection. In the future, the proposed framework is expected to further reduce the workload of 24x7 on-call experts as the LLMs and truth discovery algorithms  improve.

\vspace{-0.3cm}
\section{Results}
Our framework has been fully deployed in real-time testing to provide global earthquake event information to PAGER system for more than half a year. We have also tested our framework on a sequence of significant recent earthquake events, each denoted by a magnitude (M) on the Moment Magnitude Scale, which measures the total energy released by an earthquake. The events include the M7.2 Haiti (2021), the M7.0 Luzon, Philippines (2022), the M6.8 Luding/Sichuan, China (2022), the M6.5 Taiwan (2022), the M6.8 Michoacan, Mexico (2022), the M7.6 Papua New Guinea (2022), the M5.7 Khowy, Iran  (2022), and the M5.6 Indonesia (2022) earthquakes.

Here we present an evaluation of the performance of our framework on these real-world events. First, we present our experimental setup and performance evaluation metrics for the framework. We then characterize the data retrieved by the data pipeline, after the hierarchical event classifier, and after casualty value extraction. We also characterize the accuracy and error of the hierarchical event classifier. Finally, we evaluate the casualty estimation performance of our framework. The experimental evaluations are based on three aforementioned earthquakes: the 2021 Haiti earthquake, the 2022 Philippines earthquake, and the 2022 Luding, China earthquake, which caused substantial damage and fatalities.

\begin{figure*}[htbp]
\begin{center}
\scalebox{1}{
\begin{tabular}{c}
\includegraphics[width=2\columnwidth]{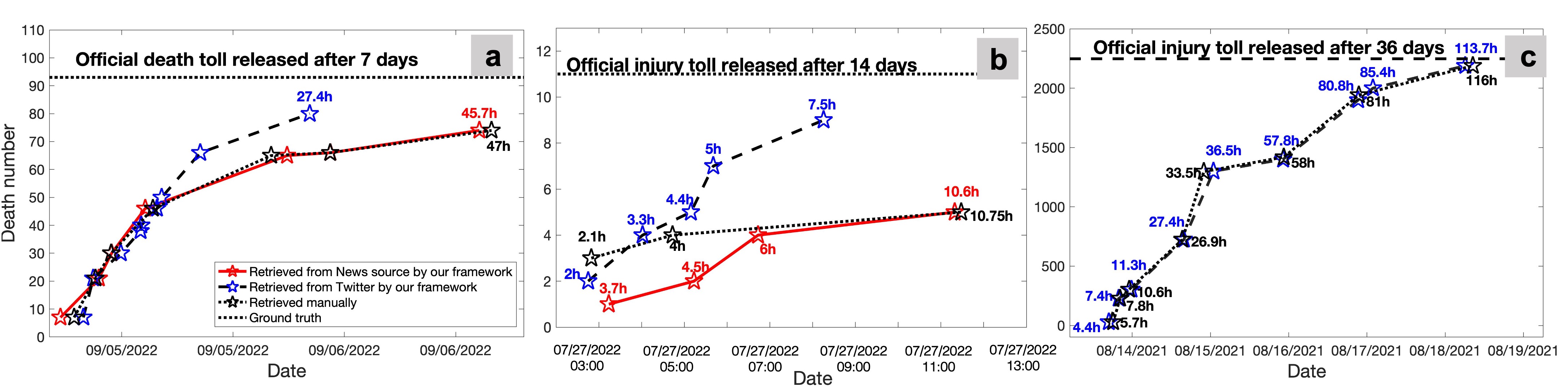}
\vspace{-0.5cm}
\end{tabular}}
\caption{A comparison between fatalities extracted from Twitter, news articles, and manually searched for (a) the 2022 Luding, China earthquake, (b) the 2022 Philippines earthquake,  (c) the 2021 Haiti earthquake as time evolves (UTC time). The solid red line represents news-sourced data and the dashed line with blue markers refers to Twitter-sourced data. The dashed line with black markers represents those manually searched. The black dotted line without any marker indicates the final official toll. Each updated fatality data point is labeled with the earliest reporting time.}
\vspace{-0.3cm}
\label{fig:nvt}
\end{center}
\end{figure*}
\noindent\emph{\textbf{Experimental Setup}}: The framework is triggered based on the magnitude of earthquake events. Our data sources include news data from News API and social media data from Twitter API (Academic research account). 
Each News API call retrieves a maximum of 100 records every half hour. Each Twitter API call retrieves up to 10,000 tweets every half hour. Twitter data provide timestamps, tweet content, user profiles, geotags, relevant news and images links, device type, and other metrics. News data are retrieved from News API (https://newsapi.org/), which covers numerous news sources and media in 14 languages from 55 countries.
The fatality data extraction process (including the backbone LLMs) is implemented using PyTorch v1.12.0 and the Docker system and conducted on a server with four NVIDIA RTX A6000 Graphics Processing Units (GPUs). 
In the real-time process, a half-hourly text data batch is fed into the information extraction model and automatically outputs deaths, injuries, city, country, if the tweet is relevant to the earthquake of interest, and which year the event occurs, as well as confidence scores associated with these answers. The results are automatically saved for dynamic truth discovery. In the truth discovery phase, we integrate all data points since the last time the fatality values are updated and finally provide the latest values as well as the first time that that value appears in the scraped text data. Finally, the casualty output data points can be fed into the PAGER loss model to update the overall fatality estimates for the earthquake.

\noindent\emph{\textbf{Performance Evaluation Metrics}}:
We characterize and evaluate our framework from two perspectives: timeliness and accuracy. Note that the final goal of this framework is to automate the fatality information extraction process to reduce the 24x7 operations expert's workload and improve the accuracy of loss estimates. To evaluate the timeliness, our goal is to achieve better or similar timeliness as manually retrieved data, for example, extracting the same casualty values earlier than manual extraction. Manually searching for casualty values is very time-consuming, requires personnel available at all hours, and depends on the agility of the expert with a wide range of search tools and social media platforms.
The accuracy is two-fold: text classification accuracy and fatality number accuracy. Because it is impossible to label every text data retrieved in our real-world experiments, we evaluate the hierarchical event classifier mainly utilizing CrisisNLP data using the accuracy rate, F1 score, and false positive rate (FPR). The FPR represents the percentage of irrelevant texts that are classified as relevant and passed to the fatality value extraction. An ideal event classification model needs to be accurate and minimize false-positive cases, as fatality information is often sensitive and critical.
Moreover, to evaluate the accuracy of the final extracted fatality value, we compare it with the officially released fatality number (often after weeks of an event).

\noindent\emph{\textbf{Retrieved Data Overview}}:
The original data retrieved from news and social media platforms mainly include a variety of languages and sources. For example, in the Luding, China, earthquake, most of the retrieved texts used Japanese, Chinese, Spanish, and English,  as shown in Figure~\ref{fig:language}(a). Whereas for the Philippines earthquake, most texts used English, Spanish, French, Japanese, Hindi, and Filipino as shown in Figure~\ref{fig:language}(b). Haiti earthquake data contain more diverse languages, dominated by English, French, Spanish, and Haitian Creole (Figure~\ref{fig:language}(c)). The distribution of languages also depicts who shares and cares about disasters or disaster-related information, combined with the Twitter account profile. We found that the majority of data are from the affected zone. In the meantime, social media accounts from earthquake-prone countries such as Japan or from neighboring countries are also actively forwarding human fatality-related information. The long-tail effect is more prominent in the Philippines event compared to that in China. This observation also helps explain why traditional methods that only focus on English text alone may not be generalized to predict disaster impacts for global earthquake events. Moreover, we found that the number of relevant texts containing casualties often increases quickly within a few hours of the event, and gradually converges as sufficient resources are allocated to the scene, as shown in Figure~\ref{fig:count}(a).

\noindent\emph{\textbf{Classified Event-Casualty-Related Text Information}}:
We evaluate the performance of our hierarchical event classifier. Our RoBERTa-based models predict accuracies of 97.4\%. As for XLM-RoBERTa, we obtain a classification accuracy of 96.7\%, a false positive rate of 0.050, casualty statistics classification accuracy of 96.1\%, and a FPR of 0.045. Although the performance slightly dips, the multilingual benefits of the XLM-RoBERTa model.  In total, there are 388 test samples for the statistic classifier and 3033 test samples for the earthquake classifier. The results show that, beyond the FPR, the two classifiers are separate and work together to effectively filter out the majority of irrelevant tweets.

We also visualize the distributions of death-related tweets, injury-related tweets, and irrelevant tweets for the three earthquakes in Figure~\ref{fig:count}. A common pattern that can be observed is that the number of related tweets increases quickly after an earthquake event occurs and gradually reduces. The rate of reduction is related to the actual death number -- usually if an earthquake cause severe human fatality, such as the Haiti earthquake that causes thousands of deaths, the number of human fatalities will be kept updated in social media for a long time (more than 5 days). Meanwhile, if the human fatality number is not significant, such as in the Philippines earthquake in 2022, the number of human fatality-related tweets shrinks quickly, as Figure~\ref{fig:count}(b) shows. 

\begin{table}[htbp]
\begin{center}
\caption{Results of different backbone LLMs for extracting death tolls from the Twitter platform for the Luding, China earthquake, compared to manual search in News platform (NA means no corresponding death number is extracted).}
\vspace{-0.3cm}
\scalebox{1.0}{
\begin{tabular}{c || c|c|c||c}
\hline
\multirow{3}{*}{Deaths}& \multicolumn{4}{|c}{Time since earthquake occurs (h)}\\\cline{2-5}
&GPT-J &GPT-Neo &  \multirow{2}{*}{BERT} &  Manual \\
&(6B params) &(1.3B params) &   &  Search \\\hline
7&	\textbf{3.0}	&3.1	&3.1&	2\\
\hline
21&	\textbf{4.1}	&6.7&	5.9&	4.3	\\
\hline
30&	\textbf{7.0}	&9.2	&8.9	&6\\
\hline
38&	\textbf{9.2}&9.3	&NA	&NA	\\
\hline
40&	\textbf{9.2}	&NA	&10.7&	NA	\\
\hline
46&	\textbf{10.9}&	11.0&	11.0&10.5	\\
\hline
50&	\textbf{11.4}&NA	&19.2	&NA 	\\
\hline
66&	\textbf{15.6}&		26.1&	31.1&29.6\\
\hline
\end{tabular}
}
\vspace{-0.5cm}
\label{sichuan_analysis}
\end{center}
\end{table}

\begin{figure*}[htbp]
\begin{center}
\scalebox{1}{
\begin{tabular}{c}
\includegraphics[width=2\columnwidth]{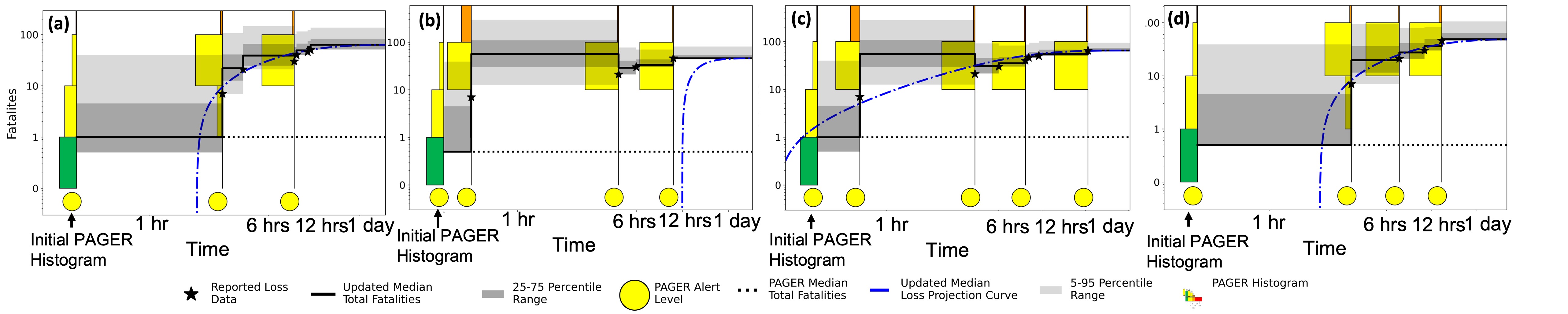}
\end{tabular}}
\vspace{-0.4cm}
\caption{PAGER fatality estimate updating using (a) Twitter data, (b) news data, (c) mixed data of Twitter and news, and (d) manually searched data}
\label{fig:pager}
\end{center}
\vspace{-0.4cm}
\end{figure*}

\noindent\emph{\textbf{Human Fatality Estimates}}:
With text data filtered by a hierarchical event classifier, we further extract the exact number of human fatalities information using LLMs and dynamic truth discovery. In this section, 

We also analyze the results of human fatality information extraction and human fatality forecasting based on the extracted human fatality information on three major earthquake events. Due to limited space, we mainly show results for the 2022 $M$6.8 Luding, China Earthquake, the 2022 M7.0 Philippines Earthquake, and the 2021 M7.2 Haiti Earthquake, and summarized the results in Figure~\ref{fig:count} and \ref{fig:nvt}.
On 04:52 September 5, 2022, UTC time, an $M$ 6.8 earthquake struck Luding County, Sichuan Province, in southwest China. A national earthquake emergency response (Level 3) was  immediately launched by the Ministry of Emergency Management of the People's Republic of China and then upgraded to Level 2 on September 6. Based on extensive field surveys, an intensity map was provided by the Ministry of Emergency Management on September 11~\cite{an2022preliminary}.
Our data collection pipeline was triggered soon after the earthquake to collect crowdsourced data from Twitter and through News API. Using our framework, the death and injury numbers were extracted as input to the PAGER loss updating the platform. We compared the timeliness of our retrieved data from the News API, Twitter, and manual search in Figure~\ref{fig:nvt}(c). It can be seen that Twitter data are updated more than either the news or our manual search data. Especially, the death numbers of $66$ and $80$ are extracted 14 hours and 20 hours earlier than the manual search.  We also compare the results of LLMs with different capacities in Table~\ref{sichuan_analysis}. It can be seen that GPT-J with 6 billion parameters more closely matches the manual search than GPT-Neo with 1.3 billion parameters and BERT. Currently, considering the limited bandwidth of deploying LLMs, GPT-J model presents a competitive capability of extracting information effectively. Meanwhile, as model parameters increase, a more powerful GPT or Open Pre-trained Transformers (OPT) model may further substantially improve our information extraction performance. Moreover, we utilize the data obtained from Twitter, news, and mixed data of both sources to update the PAGER loss estimation models, compared to the loss estimation performance using manually searched data, as shown in Figure~\ref{fig:pager}. The figures present how the forecasted probability distribution of final human casualty is updated as new data points come in. It can be seen that mixed data and news results as well as news results alone provide an estimation that the final death number will fall into the range between 10 and 100 -- which is later verified to be 93 deaths-- earliest compared to Twitter and manual search. All four types of methods provide correct forecasting because the first data point is received, demonstrating that the human fatality information retrieved by our framework can achieve comparable performance compared to manual search by human experts.

\noindent\emph{\textbf{The 2022 Philippines Earthquake}}:
The M7.0 earthquake struck the northern Philippines caused 11 deaths and 615 injuries. The recent occurrence of the earthquake makes it an ideal opportunity to experiment with actual, real-time performance of our model. 
Due to the large Tagalog and Filipino-speaking populations, we apply our XLM-RoBERTa-based hierarchical event classifier. 
Our model reports that the number of deaths will increase from 1 to 10, and the number of injuries will increase from 44 to 60 over time following the earthquake. Likewise, the official death toll was released a week later, which fell outside our time frame. 
To benchmark our social media findings, we attempted to exploit news data for official casualty reports in the 2022 Philippines earthquake.  For the news data, we treat each description as short text as a tweet, and process them similarly. As shown in Figure \ref{fig:nvt}(b), we see a notable gap between the Twitter-sourced data and many echoing data points when we source our news feeds. This gap may come from the slower speed that official outlets have compared to social media outlets. 
The M7.2 earthquake struck Haiti result in a total number of 2,248 deaths and 12,200 injured. We crawled the Twitter database for tweets containing relevant keywords or hashtags (e.g., earthquake, Haiti) to obtain the social media data for that event. 
We process each tweet with our method and obtain a time-series graph shown in Figure \ref{fig:nvt}(c). Throughout the duration, our extracted death toll rises from 29 to 2,189. We observe that our extract statistics are relatively close to the final official number but still have a slight difference because of the limited time span of the deployment. 


\vspace{-0.3cm}
\section{Conclusion}
This paper presents a novel framework for near-real-time, earthquake-induced casualty estimation from multilingual social media data. We introduce a hierarchical event classifier that categorizes and filters informative social media posts with multilingual capabilities, a process previously unexplored. We extend this casualty data extraction beyond simple categorization and directly extract relevant statistics from the tweets, including locations and uncertainties. We overcome the challenges of complex syntax and requirement-labeled data in real-time direct extraction through large language models, leveraging its capabilities of Few-Shot Learning and LLMs. We design a physical constraint-aware dynamic truth discovery model that recovers causalty estimates from massive noisy and conflicting data. In our experiments, we measure the capacity of our classification networks and evaluate the performance of our model on real-world events. Our results demonstrate that our model 
yield results that compare well with final reported losses and accurately extracted information automatically to significantly improve the timeliness and efficiency of the existing USGS PAGER system. 

\vspace{-0.3cm}
\begin{acks}
This draft manuscript is distributed solely for informational purposes. Its content is deliberative and predecisional, so it must not be disclosed or released by reviewers. Because the manuscript has not yet been approved for publication by the U.S. Geological Survey (USGS), it does not represent any official USGS finding or policy.
Any mention of commercial products is for informational purposes and does not constitute an endorsement by the U.S. government.
\end{acks}



\bibliographystyle{ACM-Reference-Format}
\bibliography{reference.bib}

\clearpage
\section*{Supplement}\label{AppendixA}
\subsection{Additional analysis for the 2022 Luding, Sichuan Earthquake}
\begin{figure*}[hbp]
\begin{center}
\scalebox{1}{
\begin{tabular}{c}
\includegraphics[width=2\columnwidth]{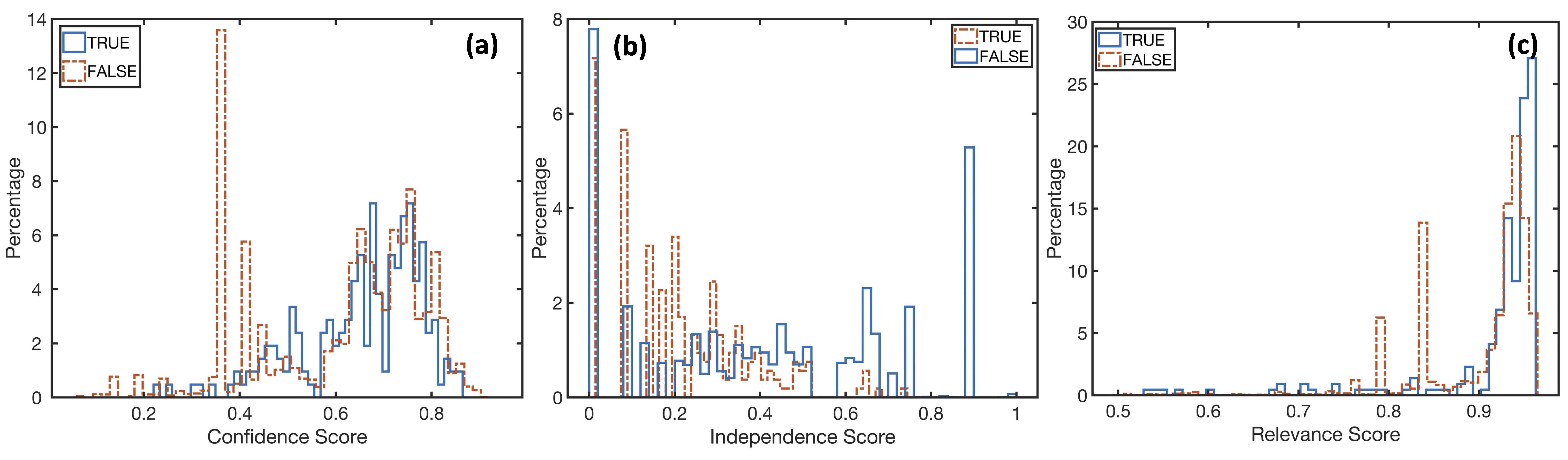}
\end{tabular}}
\caption{Confidence score, independence score, and relevance score for verified Twitter account (true, blue rectangle) and unverified Twitter account (false, orange dashed rectangle) for the filtered death tweets in the 2022 Luding, China earthquake. }
\label{fig:score}
\end{center}

\end{figure*}

\begin{figure}[hbp]
\hspace{-1mm}\begin{center}
\scalebox{0.9}{
\begin{tabular}{c}
\includegraphics[width=0.5\textwidth]{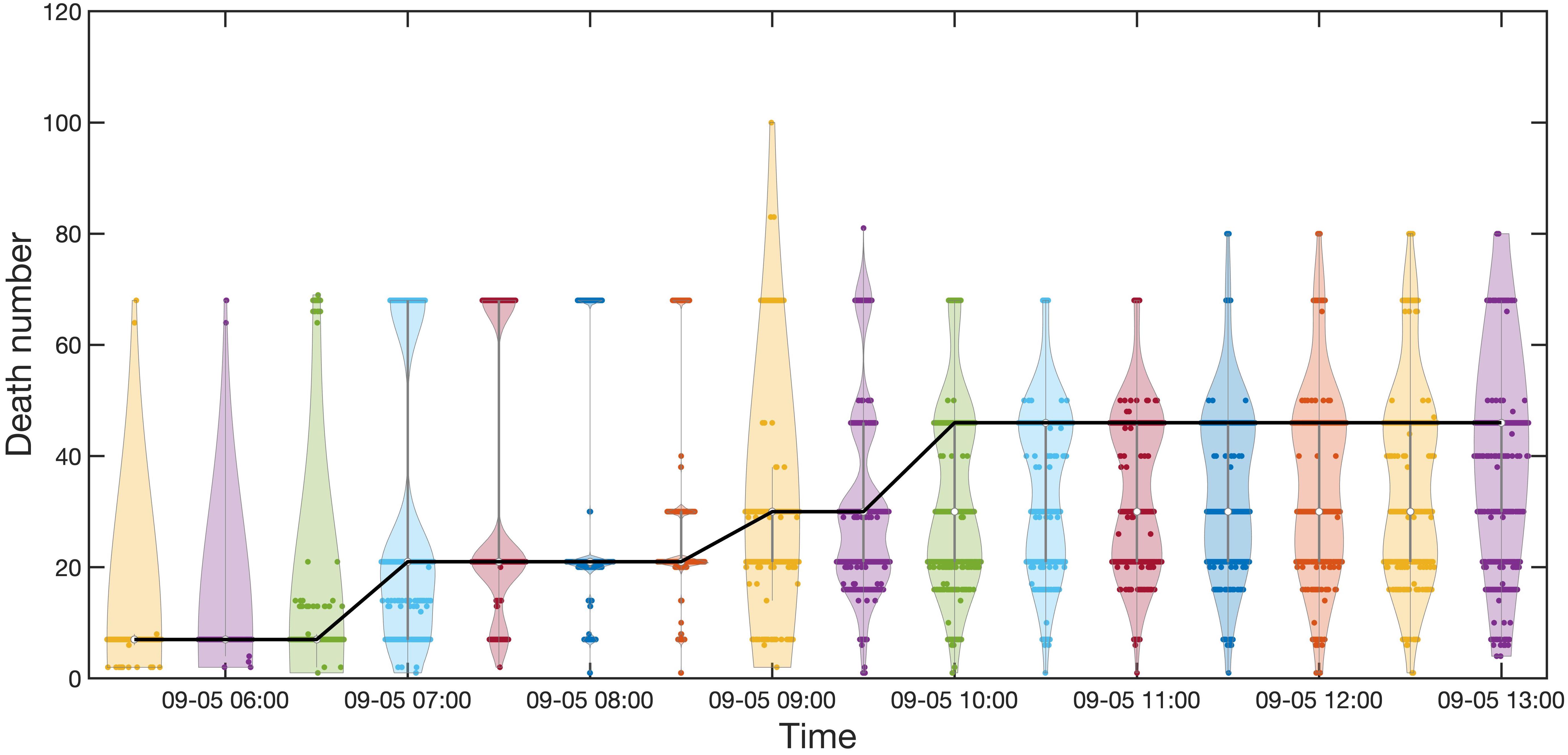}
\end{tabular}}
\vspace{-0.3cm}
\caption{Distributions of extracted human death number as time evloves.}
\label{fig:death_dist}
\end{center}
\end{figure}

Figure~\ref{fig:score} presents the independent score, confidence score, and relevance score defined in Section~\ref{dtd} to weight different source information for human fatality statistics extraction in the 2022 Luding, China earthquake. It can be seen that comparing the verified account (True) and unverified account(False), the independence score of verified account is relatively higher than unverified ones, showing that they are more independent in tweeting the information compared to unverified accounts which are mostly personal users. Similarly, the confidence score and relevance score of verified accounts are also higher than unverified ones. We also present Figure~\ref{fig:death_dist} to show the distributions of extracted death number as time evolves, as well as their mode. It can be seen that as time changes, the modes of extracted human death number increases from 7 to 46 within 9 hours. Our truth discovery algorithm aggregate these extracted death number more effectively to discover the truth earlier than simply taking the mode. For example, the death number of 21 first appears as a mode in 7 hours after the earthquake, while our algorithm discovers 21 as death number in 4.1 hours after the earthquake occurs. The results show the effectiveness and timeliness of our dynamic truth discovery combining with large language models.

\begin{table}[bp]
\begin{center}
\scalebox{1.0}{
\begin{tabular}{l || ccc}
\hline
\textbf{Model} & Acc (\%) &  F1 &  FP Rate \\
\hline
RoBERTa EC 3-Epochs
            & 97.2 & 0.97 & 0.042 \\
RoBERTa EC 4-Epochs
            & 97.4 & 0.97 & 0.034 \\ 
XLM-RoBERTa EC 3-Epochs
            & 96.7 & 0.97 & 0.050\\  
RoBERTa SC 3-Epochs
            & 95.6 & 0.96 & 0.045\\ 
RoBERTa SC 4-Epochs
            & 95.9 & 0.96 & 0.061\\ 
XLM-RoBERTa SC 4-Epochs
            & 96.1 & 0.96 & 0.045\\ 
\hline
\end{tabular}
}
\caption{Performance of the hierarchical event classifier, which integrates an earthquake classifier (EC) and a human cost statistics classifier (SC), both based on RoBERTa/XLM-RoBERTa.}
\label{c_results}
\end{center}
\vspace{-0.8cm}
\end{table}

\end{document}